\documentclass[10pt,twocolumn,letterpaper]{article}

\usepackage{iccv}
\usepackage{times}
\usepackage{epsfig}
\usepackage{graphicx}
\usepackage{amsmath}
\usepackage{amssymb}

\usepackage{multirow}
\usepackage{makecell}
\usepackage{url}
\usepackage{booktabs}

\usepackage[breaklinks=true,bookmarks=false]{hyperref}

\iccvfinalcopy 

\def\iccvPaperID{2134} 

\ificcvfinal\fi

\begin{document}

\title{Collaborative Tracking Learning for Frame-Rate-Insensitive \\ Multi-Object Tracking}

\author{Yiheng Liu,\quad  Junta Wu\thanks{First Author and Second Author contribute equally to this work.}, \quad Yi Fu \\ ByteDance Inc. \\ lyh156@mail.ustc.edu.cn, juntawu@163.com, emmafu1013@gmail.com
}

\maketitle

\ificcvfinal\thispagestyle{empty}\fi

\begin{abstract}
   Multi-object tracking (MOT) at low frame rates can reduce computational, storage and power overhead to better meet the constraints of edge devices. Many existing MOT methods suffer from significant performance degradation in low-frame-rate videos due to significant location and appearance changes between adjacent frames. To this end, we propose to explore collaborative tracking learning (ColTrack) for frame-rate-insensitive MOT in a query-based end-to-end manner. Multiple historical queries of the same target jointly track it with richer temporal descriptions. Meanwhile, we insert an information refinement module between every two temporal blocking decoders to better fuse temporal clues and refine features. Moreover, a tracking object consistency loss is proposed to guide the interaction between historical queries.
   Extensive experimental results demonstrate that in high-frame-rate videos, ColTrack obtains higher performance than state-of-the-art methods on large-scale datasets Dancetrack and BDD100K, and outperforms the existing end-to-end methods on MOT17. More importantly, ColTrack has a significant advantage over state-of-the-art methods in low-frame-rate videos, which allows it to obtain faster processing speeds by reducing frame-rate requirements while maintaining higher performance.
   Code will be released at \href{https://github.com/yolomax/ColTrack}{https://github.com/yolomax/ColTrack}
\end{abstract}

\section{Introduction}

The goal of multi-object tracking (MOT) is to estimate bounding boxes and identities of objects of interest in videos. In high-frame-rate videos, the velocities of objects are slow, which makes the difference between adjacent frames small. State-of-the-art MOT methods ~\cite{bergmann2019tracking, yu2022towards, wang2020towards, zhang2021fairmot, zhang2022bytetrack, zhou2020tracking, meinhardt2022trackformer, zeng2022motr} achieve impressive results in the high-frame-rate situation. However, limited by storage, computing, and network bandwidth, low-frame-rate videos are very common. In low-frame-rate videos, the difference between adjacent frames is larger, which degrades the performance of existing methods.

\begin{figure}

   \begin{center}
      \includegraphics[width=0.85\linewidth]{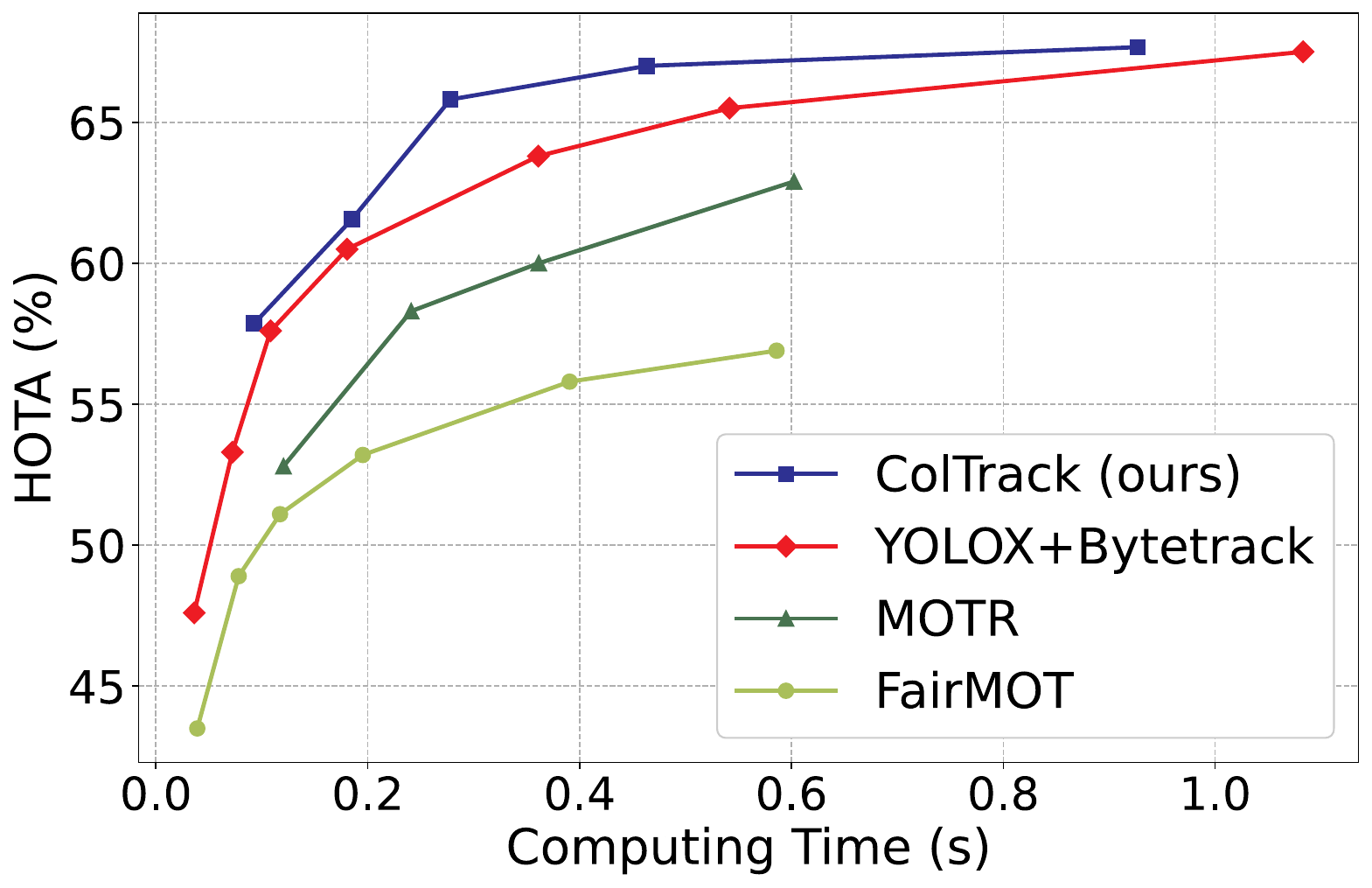}
   \end{center}
   \vspace{-0.04\linewidth}
      \caption{The comparison of the computing time for different methods to achieve the required HOTA score when tracking in a one-second video. These data are calculated based on the HOTA score and FPS of different methods on the MOT17 validation set. ColTrack still maintains high HOTA scores when tracking at low frame rates, so it achieves faster tracking speed by reducing the frame rate requirement while ensuring a high HOTA score. }
   \label{fig:eq_fps}
   \vspace{-0.03\linewidth}
   \end{figure}

The challenges caused by low-frame-rate videos are manifold. First, the displacements of objects between adjacent frames become larger, which even leads to no overlapping of the object boxes. This requires the model to match targets over a larger range, which includes more noisy objects. Furthermore, the position estimation error of the motion model (\eg the Kalman filter~\cite{kalman1960new}) is amplified and leads to significant performance degradation of the Kalman filter-based methods, \eg, Bytetrack~\cite{zhang2022bytetrack} and FairMOT~\cite{zhang2021fairmot}. 
Second, objects have severe appearance changes between adjacent frames. The viewpoints, visibilities, and poses of the objects change greatly. In addition, the sudden occlusion causes the objects to lose key appearance features rapidly. This greatly challenges some methods~\cite{zhang2021fairmot, wang2020towards} that rely on appearance features. 

Some methods focusing on the detection of emerging objects~\cite{zhou2022apptracker} or the adjustment of training strategies~\cite{feng2022towards} are proposed to improve the MOT performance at low frame rates, while these methods do not fundamentally solve the problems of unreliable features and large displacements in low-frame-rate videos. The end-to-end MOT methods~\cite{meinhardt2022trackformer, zeng2022motr} use the deformable attention-based DETR-like detection model~\cite{carion2020end, zhu2020deformable} to match objects in the current frame based on the queries from the last frame. The utilization of deformable attention allows the model to adaptively find targets in a larger range. This helps to alleviate the problem caused by the large displacements. However, the matching of objects in this way heavily depends on the quality of the queries, which cannot be guaranteed due to the unreliable features in low-frame-rate videos.

In this paper, we propose collaborative tracking learning (ColTrack) for frame-rate-insensitive MOT, which is an end-to-end MOT approach. ColTrack utilizes multiple historical queries belonging to the same object as the collaborative tracking queries to track the same target. These queries contain descriptions of the same target at different moments. Their combination effectively alleviates the impact of unreliable features. However, the introduction of multiple historical queries to track the same target is against the one-to-one matching strategy of the DETR-like detection architecture. This causes the model to not only lose the capability of inhibiting duplicate predictions, but also fail to train with the bipartite matching loss~\cite{carion2020end}. 

To address these issues, we propose an information refinement module (IRM) and insert it between every two temporal blocking decoders to enable the information fusion between collaborative tracking queries while retaining the capability of inhibiting duplicate predictions. IRM contains an information removal branch and an information addition branch to assist queries to decide how to refine themselves based on temporal clues. Furthermore, we propose a tracking object consistency loss (TOCLoss), which requires each tracking query to collect discriminative features from other historical queries for the correct tracking. The joint use of these modules enables ColTrack to achieve more stable performance at low frame rates and better track difficult targets at high frame rates.

As shown in Fig.~\ref{fig:eq_fps}, ColTrack further increases the processing speed of the video by reducing the frame rate requirement while ensuring higher accuracy. In contrast, existing methods~\cite{zhang2022bytetrack, zhang2021fairmot, zeng2022motr} require higher frame rates to achieve high accuracy, which results in more video frames being processed and lower processing speed. 

To summarize, our contributions are as follows:
\begin{itemize} \setlength{\itemsep}{0pt}
   \item We propose a query-based end-to-end model ColTrack that uses the collaborative tracking of multiple historical queries to achieve stable performance even at low frame rates. 
   \item We further devise a IRM module to allow each query to better fuse information based on temporal cues. The proposed TOCLoss guides queries to collect valuable clues from other historical queries.
   \item ColTrack not only outperforms state-of-the-art methods on large-scale datasets under high frame rates but also achieves higher and more stable performance under low frame rates. This allows it to obtain a higher equivalent FPS by reducing the frame rate requirement.
\end{itemize}


\section{Related Works}\label{sec:relatedWrok}

\textbf{Classical MOT Methods.} Most classical MOT methods follow the tracking-by-detection paradigm by detecting the object-bounding boxes first and then tracking objects by data association. For example, SORT~\cite{bewley2016simple} , DeepSORT~\cite{wojke2017simple}, and ByteTrack~\cite{zhang2022bytetrack} all follow this paradigm. They use Kalman filters to model tracks and update the underlying locations or features at each time step. JDE~\cite{wang2020towards}, FairMOT~\cite{zhang2021fairmot}, and Unicorn~\cite{yan2022towards} further explore the MOT system that jointly learns object detection and appearance embedding with a shared model.

\textbf{Transformer-Based MOT Methods.} Recently, the transformer has been applied in various computer vision tasks and achieved great success. TransTrack~\cite{sun2020transtrack} introduces a query-key mechanism based on transformer architecture. It uses object features from the last frame as queries and tracks existing targets by associating bounding box locations. Trackformer~\cite{meinhardt2022trackformer} and MOTR~\cite{zeng2022motr} follow the DETR structure and both introduce autoregressive track queries to the transformer decoder to achieve implicit data association between frames. TransMOT~\cite{chu2023transmot} augments the transformer with spatial-temporal graphs to enhance the modeling capabilities of spatial relationships, which builds a new tracking-by-attention paradigm to MOT.

\begin{figure*}
   \begin{center}
      \includegraphics[width=0.99\linewidth]{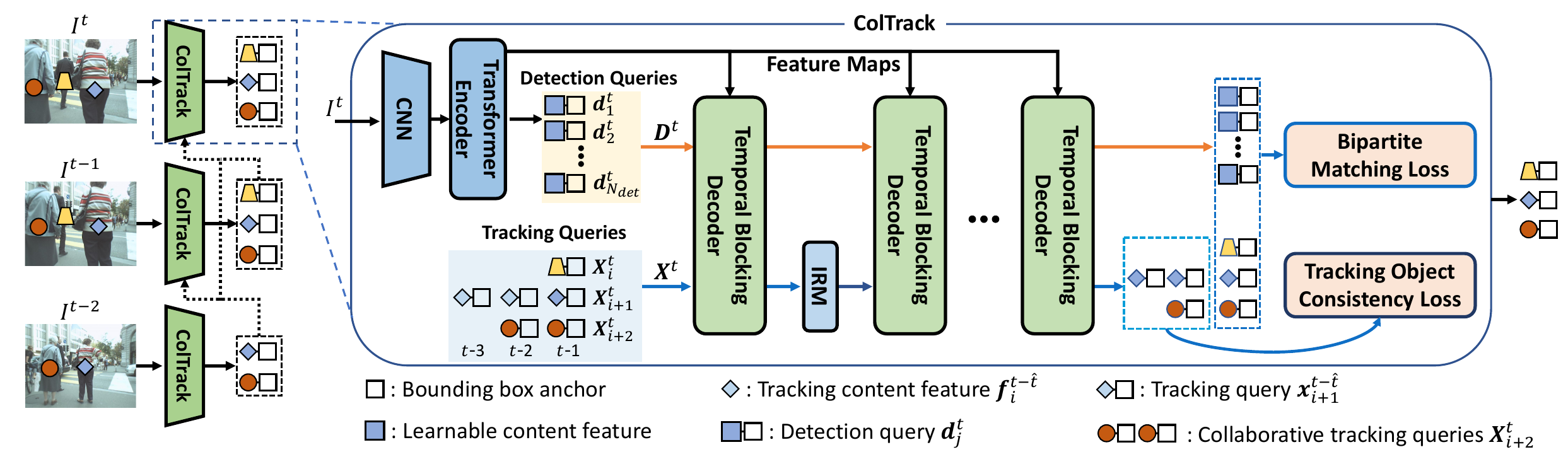}
   \end{center}
   \vspace{-0.015\linewidth}
      \caption{The overall workflow of ColTrack. The transformer encoder provides image feature maps and detection queries of emerging objects. Multiple historical queries of each tracked object constitute its collaborative tracking queries for the joint tracking of it.  The combined queries are fed into multiple temporal blocking decoders to iteratively refine the predictions. An information refinement module (IRM) is inserted between every two decoders for collaborative tracking queries belonging to the same target to integrate temporal clues and refine themselves. The tracking object consistency loss guides the consistent tracking of historical queries to the corresponding target.}
   \label{fig:arc:overview}
   \vspace{-0.01\linewidth}
   \end{figure*}

\textbf{MOT Methods with Historical Features.}
Temporal information is crucial for MOT as it is essentially a video analysis task. MTrack~\cite{yu2022towards} extracts discriminative representation to track objects in occlusion scenarios. It obtains the weighted feature representation of the trajectory according to the cosine similarity of historical features. GTR~\cite{zhou2022global} matches current detection results with tracked objects by calculating the similarity between current features and multiple historical features. The interaction between historical features in these methods is limited, and the complementary information between features is not fully exploited. 

MeMOT~\cite{cai2022memot} designs the instance feature memory banks to generate a better track query for each object. However, fusing information from multiple historical queries into one query through only one module inevitably leads to information loss. To avoid these problems, we introduce multiple historical queries as collaborative tracking queries to jointly participate in the tracking. We allow multiple interactions of historical queries to fully integrate information. 

\textbf{MOT at Low Frame Rates.}
The tracking in low-frame-rate videos is a big challenge for the MOT task due to the large difference between adjacent frames. 
APPTracker \cite{zhou2022apptracker} adds an appear predictor (APP) head into CenterTrack \cite{zhou2020tracking} architecture to detect objects that newly appear in the current frame. Since CenterTrack \cite{zhou2020tracking} cannot predict the correct displacements for objects having visibility flips across frames, APP improves the performance of CenterTrack at low frame rates. However, APPTracker focuses more on the detection of new objects, which limits its performance in low-frame-rate videos. 

FraMOT~\cite{feng2022towards} directly introduces frame rate cues to the association module to handle the low-frame-rate case and applies tracking patterns to reduce the gap between the training phase and the inference one. However, the tracking patterns need to be updated periodically, which severely slows down the training. Besides, the frame rate information can only provide coarse clues for matching, which makes it unable to fundamentally solve the problems caused by a low frame rate.

Different from these methods, our method is specifically devised to address the large appearance and location change problem when tracking at a low frame rate. This enables our method to achieve satisfactory performance in both high-frame-rate and low-frame-rate videos.

\section{Our Method}\label{sec:ourMethod}

 \subsection{Overview}

 As shown in Fig.~\ref{fig:arc:overview}, the proposed collaborative tracking learning model (ColTrack) for frame-rate-insensitive MOT is built on the encoder-decoder Transformer \cite{zhang2022dino} architecture. 
 Given a sequence of video frames \(\{ \mathbf{I}^{1}, \mathbf{I}^{2}, \cdots, \mathbf{I}^{T} \}\), 
 ColTrack tracks objects of interest in each frame and predicts their class and bounding boxes. For the video frame \(\mathbf{I}^{t}\), the CNN model extracts its features, which are then fed to the transformer encoder to provide feature maps and \(N_{\mathrm{det}}\) candidate target anchors. Each anchor \(\hat{\mathbf{b}}^{t}_j\) is the predicted bounding box of one object. These anchors together with a set of learnable content features constitute \(N_{\mathrm{det}}\) detection queries $ \mathbf{D}^t = \{\mathbf{d}^t_j\}_{j=1:{N_{\mathrm{det}}}}$. Detection queries are used to detect new objects appearing in the current frame.

 Similar to the existing methods~\cite{meinhardt2022trackformer, zeng2022motr, cai2022memot}, ColTrack uses the queries output from previous video frames as the tracking queries to track the tracked targets. But unlike these methods that only construct one tracking query for each tracked target, ColTrack utilizes multiple historical features of each tracked target to construct collaborative tracking queries for collaborative tracking in the current frame. The tracking queries \( \mathbf{X}^t \) and detection queries \(\mathbf{D}^t\) are combined and fed to subsequent multiple temporal blocking decoders to iteratively refine features and bounding boxes. An information refinement module (IRM) is inserted between two adjacent decoders to allow information fusion of historical features belonging to the same target. 
 
 The output queries of each decoder are sent to the bipartite matching loss and the tracking object consistency loss to guide the training of the model (in Fig.~\ref{fig:arc:overview}, the losses of the middle decoders are not drawn for brevity). The queries of the matched targets output by the last decoder are sent to subsequent frames as new historical queries. After sequentially processing each frame, ColTrack obtains the tracking results of the entire video.
 
 \subsection{Collaborative Tracking Queries}

 To enhance the model's tracking at a low frame rate, we propose a collaborative tracking method based on historical features. For the ${i}^\mathrm{th}$ tracked object, we store its historical features \( \mathbf{F}^{t} = \{\mathbf{f}^{t-\hat{t}}_i\}_{\hat{t}=1:N_i^t}, N_i^t \leq N_{\mathrm{max}} \). \(N_{\mathrm{max}}\) is the max memory size and \(N_i^t\) is the number of historical features. \(\mathbf{f}^{t-\hat{t}}_i  \in {\mathbb{R}}^{d}  \) is the output content feature of the corresponding target in ${t-\hat{t}}$ frame. 
 Considering that usually the latest location prediction \(\hat{\mathbf{b}}^{t-1}_i\) is closer to the location of the target in the current frame than \(\{\hat{\mathbf{b}}^{t-\hat{t}}_i\}_{\hat{t}=2}^{N_i^t}\), we adopt \(\hat{\mathbf{b}}^{t-1}_i\) as the target anchor, which is combined with each historical feature \(\mathbf{f}^{t-\hat{t}}_i\) to construct the tracking query \(\mathbf{x}^{t-\hat{t}}_i\). 
 The \(N_i^t\) queries \( \mathbf{X}^t_i = \{\mathbf{x}^{t-\hat{t}}_i\}_{\hat{t}=1:N_i^t} \) are considered as collaborative tracking queries and jointly track the same target in the current frame. They are initialized with different historical features and the same anchor \(\hat{\mathbf{b}}^{t-1}_i\). 
 
 The benefits of the introduction of collaborative tracking queries are manifold. First, this approach provides more abundant descriptions of objects to alleviate the impact of unreliable features in low-frame-rate videos. Multiple historical queries containing different temporal clues directly participate in the tracking of the target in the current frame. During the iterative refinement of multiple decoders and IRM modules, each historical query adaptively collects valuable clues based on the tracking results of other historical queries to obtain a more accurate feature description for better tracking in the next decoder. The direct participation of historical queries enables the contained valuable information to be mined better.
 
 Second, this approach is more conducive to training a frame-rate-insensitive MOT model. ColTrack requires historical features of the same target with different time spans from the current frame to track the target correctly. This is equivalent to training the model to track targets at different frame rates. Older features usually have a larger appearance difference from the current target. This requires ColTrack to be able to extract more robust features, and the refinement module IRM should be able to better integrate effective information from other historical features to deal with the impact of frame rate.

 \begin{figure}
   \begin{center}
      \includegraphics[width=0.95\linewidth]{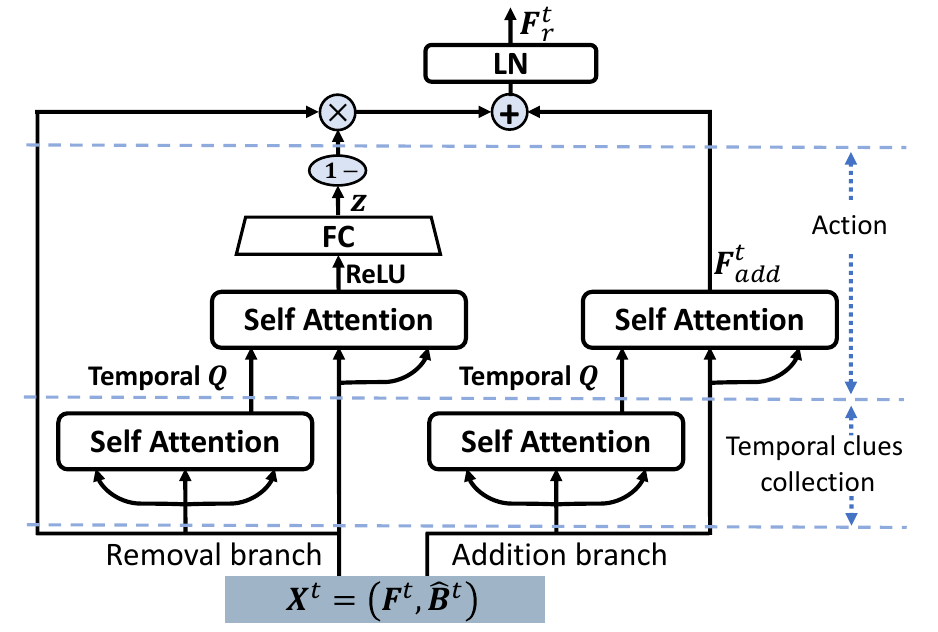}
   \end{center}
   \vspace{-0.03\linewidth}
      \caption{The architecture of the information refinement module (IRM). It mainly includes two branches: the information removal branch and the information addition branch. Each branch is mainly composed of temporal clues collection layer and action module. }
   \label{fig:arc:irm}
   \vspace{-0.03\linewidth}
   \end{figure}

 \subsection{Information Refinement Module}

 As demonstrated in \cite{carion2020end}, benefit from the self-attention mechanism over the activations in decoders, the DETR-like detection model~\cite{carion2020end, zhu2020deformable, zhang2022dino} can discard the non-maximum suppression (NMS) post-processing. This means that the decoder suppresses the case where multiple queries detect the same object to inhibit duplicate predictions. But in our method, collaborative tracking queries of the same object have to track the same object, which is against the design of the decoder. To address this issue, we propose temporal blocking decoders, which avoid the interaction between historical tracking queries of the same target by modifying the attention mask of the multi-head self-attention module in the decoder.

 The temporal interaction blocking capability provided by the temporal blocking decoder avoids mutual inhibition between collaborative tracking queries of the same target, but it also makes it impossible to exchange information between collaborative tracking queries to refine themselves. To solve this problem, we devise a new information refinement module (IRM) to allow the interaction between collaborative tracking queries and refine them. As shown in Fig.~\ref{fig:arc:irm}, the multi-head self-attention layer is the main component of the IRM. The input of IRM is the \(N^t\) collaborative tracking queries of all the tracked objects, where \(N^t=\sum_i{N^t_i}\). The output is the refined features. We modify the attention mask to avoid the interaction of features from different objects. 
 
 Since the main contribution of IRM is to allow collaborative tracking queries belonging to the same target to interact and update information, then for each query, it is necessary to decide how to remove old information and what new information to add. Inspired by this, the IRM we devised mainly consists of two branches: an information removal branch and an information addition branch. Each branch contains two parts, \emph{i.e.} the temporal clues collection part and the action part.

A single multi-head self-attention layer of the transformer is not able to compute any cross-correlations between the queries~\cite{carion2020end}. This is because a single self-attention layer can only organize information once based on the similarities of features. The model cannot see the global information to decide how to output. Therefore, we add a multi-head self-attention module as the temporal clues collection part to collect global temporal clues for each tracking query. 
 
 For the information addition branch, the action part uses the collected global temporal clues as queries and uses the original content features as keys and values to generate the content features \(\mathbf{F}^{t}_{\mathrm{add}} \in {\mathbb{R}}^{N^t \times d} \) that needs to be added. Compared with the information addition branch, the information removal branch has one more fully connected layer followed by a sigmoid layer to map the embedding to a  gating vector \(\mathbf{z} \in {\mathbb{R}}^{N^t \times d_{\mathrm{head}}} \). \(d_{\mathrm{head}}\) is the head number of the multi-head self-attention module. We divide each content feature into \(d_{\mathrm{head}}\) groups, and use the gating vector \(\mathbf{z}\) to control the degree of deletion of each group of features. Then, we formulate the refinement of content features as 
 \begin{equation}
   \label{equ:query:refine}
   \mathbf{F}^{t}_r = \mathrm{LN}(2\mathbf{F}^{t} \times (1 - \mathbf{z} ) + \mathbf{F}^{t}_{\mathrm{add}} ) \,,
\end{equation}
where \(\mathrm{LN}(\cdot)\) is the layer normalization~\cite{ba2016layer}. \(\mathbf{F}^{t}_r \in {\mathbb{R}}^{N^t \times d} \) is the refined content features. \(\mathbf{F}^{t} \times (1 - \mathbf{z} )\) denotes the reserved features. We double it to increase its weight. The refined content features \(\mathbf{F}^{t}_r \) are combined with the corresponding anchors to form the refined tracking queries, which are sent to the next temporal blocking decoder. The combination of the temporal blocking decoders and IRM avoids the duplicate predictions problem and ensures effective information interaction between historical tracking queries belonging to the same target.

Besides, we insert IRM between every two decoders, which allows multiple interactions between historical queries. After each decoder, each query completes a target detection in the current frame and obtains new features and position estimation. During multiple interactions through IRMs, collaborative tracking queries continuously exchange varied new observations to obtain a more comprehensive description of the target. Collaborative tracking based on multiple information interactions is very important for stable tracking at a low frame rate.

 \subsection{Tracking Object Consistency Loss}

 In the existing transformer-based end-to-end MOT methods~\cite{zeng2022motr, cai2022memot, meinhardt2022trackformer}, the bipartite matching loss \cite{carion2020end} is adopted to train the network. 
 In our method, for each tracked target, due to the introduction of collaborative tracking queries, predictions of multiple queries belonging to the same target are matched with the same ground truth. 
 This prevents us from directly using the one-to-one matching strategy to calculate the bipartite matching loss.
 Therefore, we propose a tracking object consistency loss (TOCLoss) to handle the training of the collaborative tracking queries. Then, the bipartite matching loss \( \mathcal{L}^t_{\mathrm{bip}} \) and the TOCLoss \(\mathcal{L}^t_{\mathrm{toc}}\) make up the overall training objective \(\mathcal{L}\), which is formulated as 
 \begin{equation}
   \label{equ:loss:final}
   \mathcal{L} = \sum\nolimits_{t=1}^{T}(\mathcal{L}^t_{\mathrm{bip}} + \mathcal{L}^t_{\mathrm{toc}}) \,.
\end{equation}

 For collaborative tracking queries \( \mathbf{X}^t_i = \{\mathbf{x}^{t-\hat{t}}_i\}_{\hat{t}=1:N_i^t} \) of one target, we use \( \{\hat{\mathbf{y}}^{t-\hat{t}}_i\}_{\hat{t}=1:N_i^t} \) to denote their predictions in the current frame. Each prediction \(\hat{\mathbf{y}}^{t-\hat{t}}_i\) contains the predicted class probabilities \(\hat{\mathbf{p}}^{t-\hat{t}}_i\) and box prediction \(\hat{\mathbf{b}}^{t-\hat{t}}_i\). We use \( \psi^t \) to denote the identity set of all tracked objects.
 
 As shown in Fig.~\ref{fig:arc:overview}, the tracking predictions \( \{ \hat{\mathbf{y}}^{t-1}_i \}_{i \in \psi^t }\) of the latest tracking queries of all tracks are combined with the predictions of the detection queries to participate in the bipartite matching. The mapping \( \pi^t \) between predictions and ground truth objects is determined either via track identities or costs based on object class and bounding box similarity \cite{meinhardt2022trackformer}.
  Then, we calculate the bipartite matching loss for them by mapping \( \pi^t \), which is denoted as \( \mathcal{L}^t_{\mathrm{bip}} \).
 Since the additional predictions \( \{\hat{\mathbf{y}}^{t-\hat{t}}_i\}_{\hat{t}=2:N_i^t} \) of each track are separated from \( \mathcal{L}^t_{\mathrm{bip}} \), \( \mathcal{L}^t_{\mathrm{bip}} \) satisfies the one-to-one matching requirement. 

 As for the remaining predictions \( \{\hat{\mathbf{y}}^{t-\hat{t}}_i\}_{\hat{t}=2:N_i^t} \) of the $i^\mathrm{th}$ track, since they and \(\hat{\mathbf{y}}^{t-1}_i\) have the same identity, they can share the mapping \( \pi^t \) between them and the ground truth objects. Then, we define the loss of each prediction as 
 \begin{small}
 \begin{equation}
   \label{equ:loss:history}
   \mathcal{L}^t_i(\hat{\mathbf{y}}^{t-\hat{t}}_i, \pi^t) \!\! = \!\! \left\{
   \begin{array}{l}
   \!\!\!\! -\log{\hat{\mathbf{p}}^{t-\hat{t}}_i}(\pi^t(i)) \! + \! \mathcal{L}_{\mathrm{b}}(\hat{\mathbf{b}}^{t-\hat{t}}_i, \pi^t) \,\,\,\,\,\, \mathrm{if} \,  i \in \pi^t   \,, \\
   \!\!\!\! -\log{\hat{\mathbf{p}}^{t-\hat{t}}_i}(0) \quad \quad \quad \quad \quad \quad \quad  \   \,\,\,\,\,\,   \mathrm{otherwise} \,,
   \end{array} \right. 
   \end{equation}
\end{small}
where \(\hat{\mathbf{p}}^{t-\hat{t}}_i(\pi^t(i))\) is the predicted probability of the assigned class obtained from mapping \( \pi^t \). \(\hat{\mathbf{p}}^{t-\hat{t}}_i(0)\) is probability of background class. \(\mathcal{L}_{\mathrm{b}}(\hat{\mathbf{b}}^{t-\hat{t}}_i, \pi^t)\) is the bounding box loss~\cite{zhang2022dino}. Then, the tracking object consistency loss \( \mathcal{L}^t_{\mathrm{toc}} \) of the remaining historical tracking queries is formulated as
\begin{equation}
   \label{equ:loss:allhistory}
   \mathcal{L}^t_{\mathrm{toc}} = \frac{ \sum_{i \in \psi^t}\sum\nolimits_{\hat{t}=2}^{N^t_i} \mathcal{L}^t_i(\hat{\mathbf{y}}^{t-\hat{t}}_i, \pi^t)  }{ \sum\nolimits_{t=1}^{T}N^t_{\mathrm{his}}  } \,,
\end{equation}
where \(N^t_{\mathrm{his}}\) is the number of queries that are assigned ground truth objects by mapping \(\pi^t(i)\) among the remaining historical tracking queries, which is formulated as 
\begin{equation}
   \label{equ:loss:historynum}
   N^t_{\mathrm{his}} = \sum_{i \in \pi^t} \lvert  \{\hat{\mathbf{y}}^{t-\hat{t}}_i\}_{\hat{t}=2:N_i^t}  \rvert \,.
\end{equation}

In the inference stage, for a track, given the predictions of all collaborative tracking queries, only the output \(\hat{\mathbf{y}}^{t-1}_i \) of query \(\mathbf{x}^{t-1}_i\) is adopted as the predicted location and score of the target in the current frame. The remaining historical queries provide temporal cues to assist tracking and their final output is ignored. We consider the object to appear when the score is greater than a threshold \(\sigma\), otherwise, the object is lost. The queries of lost targets are kept for up to \(N_\mathrm{keep}\) frames.

\begin{figure*}[t]
   \begin{center}
   \includegraphics[width=0.315\linewidth]{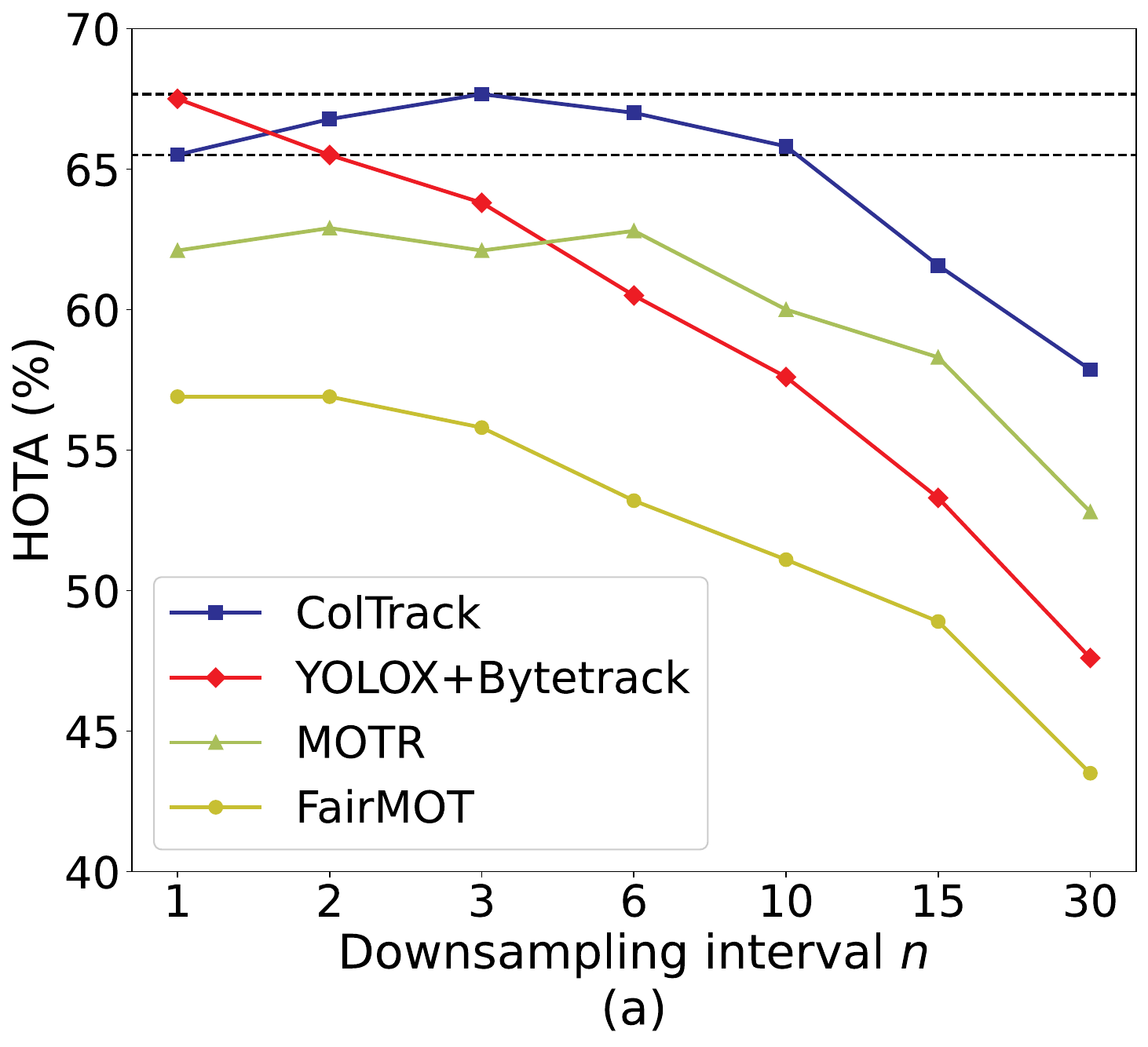}
   \hspace{0.25cm}
   \includegraphics[width=0.315\linewidth]{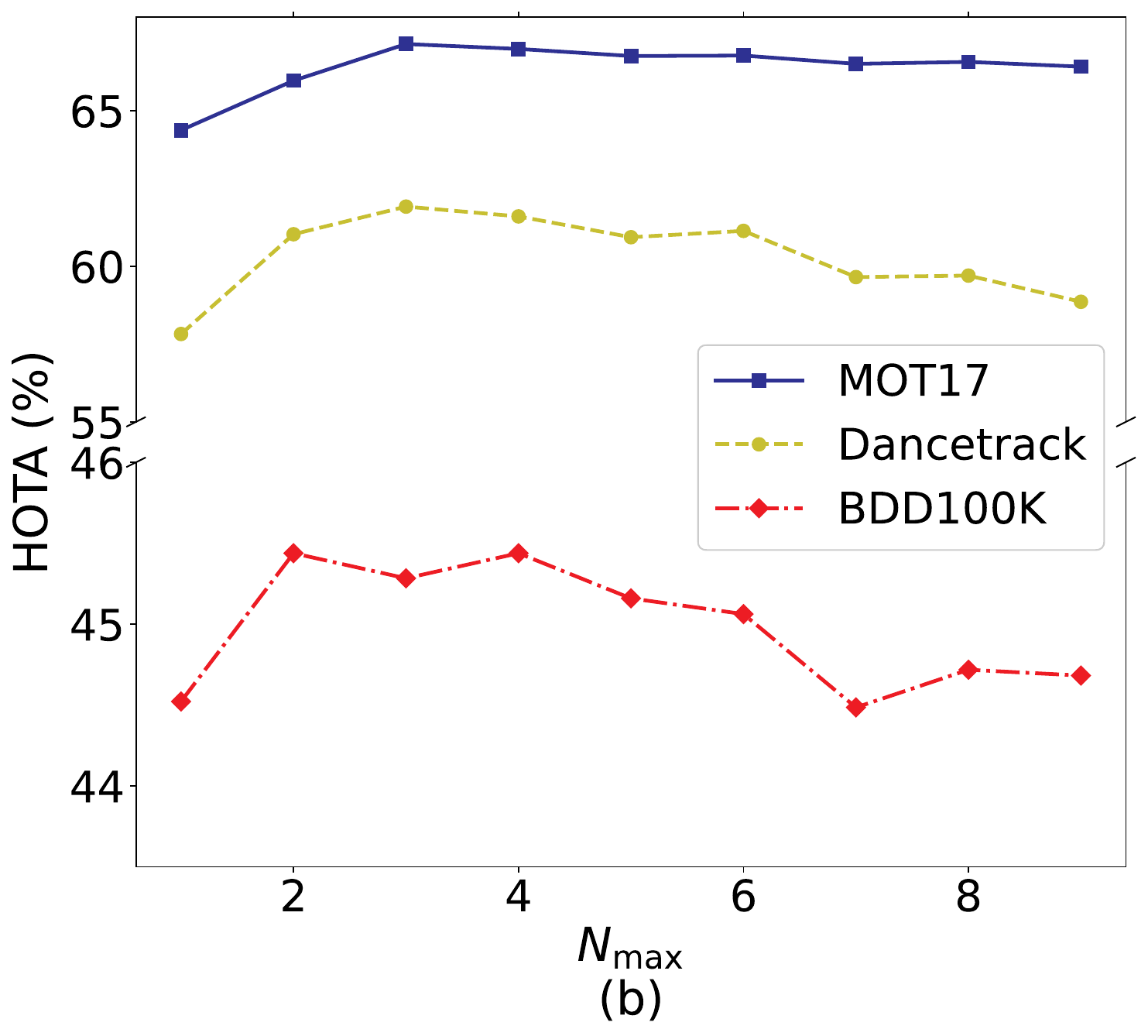}
   \hspace{0.25cm}
   \includegraphics[width=0.315\linewidth]{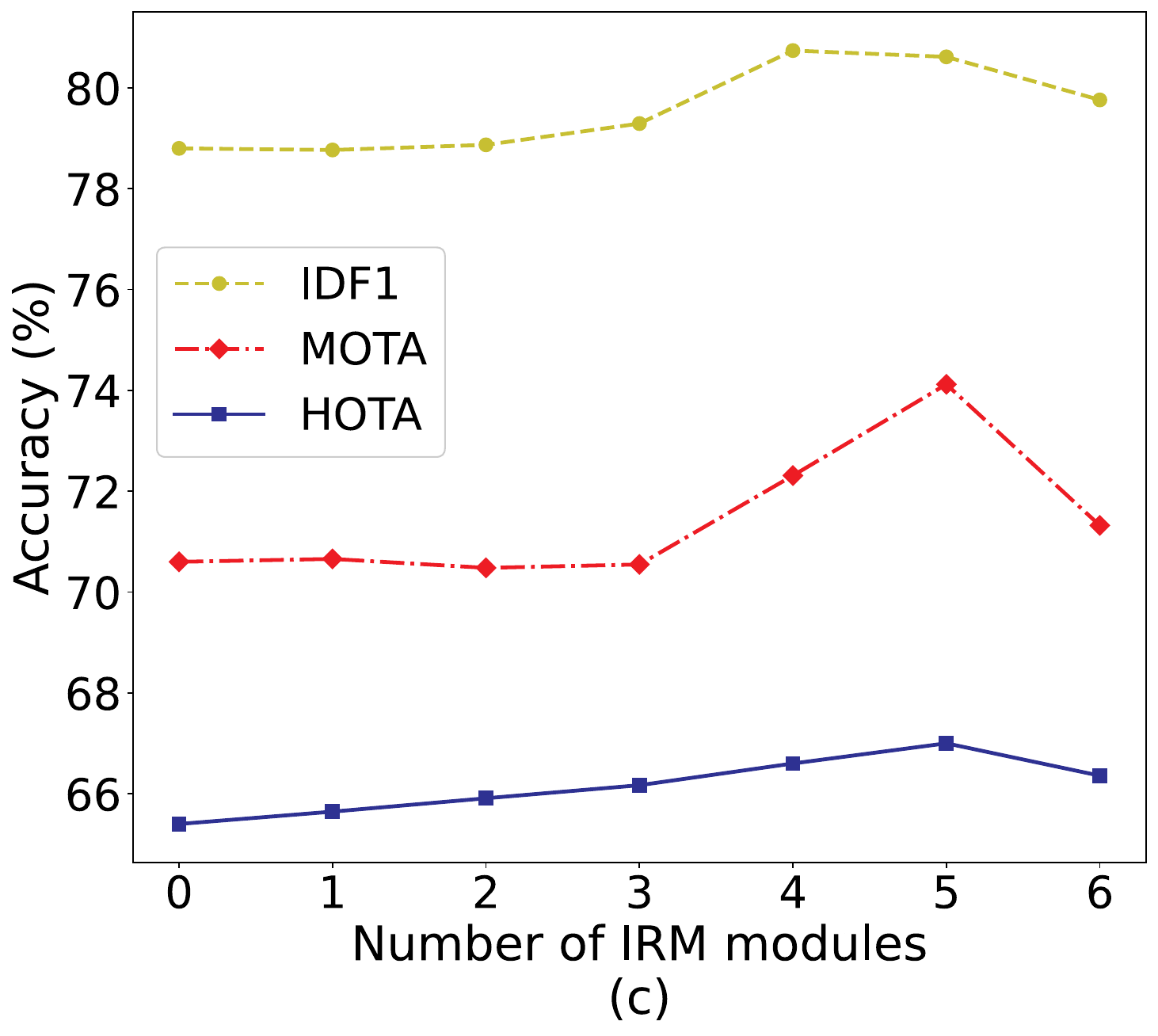}
   \end{center}
   \vspace{-0.02\linewidth}
   \caption{ (a) Performance comparison of ColTrack with existing methods on videos at different frame rates on the validation set of MOT17. \(n\) represents that the videos are downsampled at a sampling interval of \(n\). (b) Performance of ColTrack under different \(N_{\mathrm{max}}\) in low-frame-rate videos (\(n=6\)). \(N_{\mathrm{max}}\) is the maximum number of collaborative historical queries of each track. (c) Performances of ColTrack with different numbers of IRMs on the validation set of MOT17 under low frame rate (\(n=6\)).}
   \label{fig:abla:method}
   \vspace{-0.01\linewidth}
   \end{figure*}

\section{Experiment}\label{sec:experiments}
   
   \subsection{Experimental Setup}
   
   \textbf{Datasets.} We evaluate our method on three popular multi-object tracking datasets, including MOT17~\cite{dendorfer2021motchallenge},  Dancetrack~\cite{sun2022dancetrack} and BDD100K~\cite{yu2020BDD100K}. 
   We use the private detection protocol for MOT17. Following ~\cite{zhang2022bytetrack, zhang2021fairmot, zeng2022motr}, the CrowdHuman dataset~\cite{shao2018crowdhuman} is added to build the joint dataset when training on the MOT17 and Dancetrack. MOT17 does not provide a validation set. In the ablation studies part, we use the first half of each video in the training set for training and the last half for validation following~\cite{zhang2022bytetrack, zhang2021fairmot, zeng2022motr}. 
   
   \textbf{Evaluation Protocols.} We follow the standard evaluation protocols including CLEAR metrics (MOTA, IDs)~\cite{bernardin2008evaluating}, Identity F1 Score (IDF1)~\cite{ristani2016performance}, and HOTA~\cite{luiten2021hota}. MOTA focuses more on the detection performance, while IDF1 focuses more on the association performance. HOTA~\cite{luiten2021hota} balances the impact of detection and data association well. Therefore, we take HOTA as the main metric.

   \textbf{Implementation Details.} ColTrack extends the DETR-like deformable detection model DINO~\cite{zhang2022dino} and takes ResNet50~\cite{he2016deep} as the CNN feature extractor. The tracking-by-detection version of the baseline method is represented as Baseline+Bytetrack, which directly trains a detection model and uses Bytetrack~\cite{zhang2022bytetrack} to associate objects. The model is trained for 40 epochs. The end-to-end (E2E) version of the baseline method is denoted as Baseline+E2E, which directly takes the tracking results of the last frame as the tracking queries~\cite{meinhardt2022trackformer, zeng2022motr} in the current frame. To reduce the GPU memory usage and increase the video clip length during training, we use the detection model trained by Baseline+Bytetrack to initialize the CNN and encoders of Baseline+E2E and ColTrack, whose parameters are frozen. The decoders and 300 learnable query embeddings are trained from scratch. The model is trained 60 epochs on MOT17, 40 epochs on Dancetrack and 20 epochs on BDD100K. 
   
   Following~\cite{zhang2022bytetrack}, the input frames are resized to \(1440 \times 800\). The data augmentation includes multi-scale training, Mosaic \cite{bochkovskiy2020yolov4} and Mixup \cite{zhang2017mixup}. All the models are trained by AdamW algorithm with an initial learning rate of \(1 \times 10^{-4}\) and weight decay of \(1 \times 10^{-4}\). The learning rate is scaled by \(\times0.1\) during the last 10 epochs. The batch size is 8 video clips and each has 4 frames. \(N_{\mathrm{det}}=300, d_{\mathrm{head}}=8, \sigma=0.6, N_\mathrm{keep}=5, N_{\mathrm{max}}=3 \). Following~\cite{zhang2022bytetrack}, we measure FPS with FP16-precision and batch size of 1 on a single V100 GPU.
 
   To obtain videos under different frame rates, following~\cite{zhou2022apptracker}, we sample frames at a fixed interval \(n\) from the original dataset to obtain low-frame-rate videos. \(n\) varies from 1 to 30. The larger \(n\), the lower the frame rate, and the larger difference between adjacent frames.

   \subsection{Effect of Frame Rate on MOT}

   \newcommand{\tabincell}[2]{\begin{tabular}{@{}#1@{}}#2\end{tabular}}  

      \begin{table}
         \small
         \begin{center}
            \setlength{\tabcolsep}{1.3mm}{
         \begin{tabular}{lcccc}
         \hline
          & ColTrack & \tabincell{c}{YOLOX+\\Bytetrack} & FairMOT  & MOTR \\
         \hline
         FPS\(^*\)          & 10.8 & 27.7 & 25.6  & 8.3  \\
         FPS\(^\mathrm{eq}\) & 32.4 & 27.7 & 25.6  & 16.6 \\
         HOTA               & 67.7 & 67.5 & 56.9 & 62.9 \\
         \hline
         \end{tabular}}
         \end{center}
         \caption{FPS and the highest HOTA score comparison of different methods on the validation set of MOT17.  FPS\(^*\) denotes the FPS we obtained by reproducing the methods on the same machine. FPS\(^\mathrm{eq}\) is the equivalent FPS, which is obtained by multiplying FPS\(^*\) by the downsampling interval \(n\) when each method obtains the highest HOTA score. The corresponding maximum HOTA score is also listed.}
         \label{tab:vel:fps}
         \vspace{-0.02\linewidth}
         \end{table}

   \textbf{Performance at low frame rates}. 
   In Fig.~\ref{fig:abla:method}(a), we compare the performance of ColTrack with existing methods~\cite{zhang2022bytetrack, zhang2021fairmot, zeng2022motr} at different frame rates on MOT17 dataset. 
   
   MOTR~\cite{zeng2022motr} has stable HOTA when \(n\) is less than 6. The HOTA of ColTrack is stable when \(n\) is varied from 1 to 10. This is benefited from the deformable attention-based end-to-end architecture, which relies on content features to track the target. This approach has more advantageous in a low-frame-rate situation that has large displacements. 
   
   Compared to MOTR, the HOTA performance and frame rate robustness of ColTrack are significantly better, which benefits from the introduction of collaborative tracking queries and the exquisite design of each module. They provide more temporal cues for the model to obtain more comprehensive and accurate descriptions of objects. 

   For YOLOX+Bytetrack~\cite{zhang2022bytetrack} and ReID-based FairMOT~\cite{zhang2021fairmot}, the ablation models use the officially provided weight. Both of them show a rapid performance drop when the frame rate is lowered. This is because these methods all rely on the Kalman filter and IOU matching, which are unreliable for fast-moving objects. 
   
   \textbf{Equivalent FPS}.
   The HOTA accuracy of ColTrack when \(n=3\) is higher than that of YOLOX+Bytetrack when \(n=1\). Therefore, as shown in Table.~\ref{tab:vel:fps}, although ColTrack doesn't have the highest FPS, it achieves a higher equivalent FPS by reducing the frame rate requirement. This allows it to process a video in a shorter time while ensuring high accuracy. Baseline+E2E has an 11.3 FPS running
   speed. Comparing the FPS of Baseline+E2E and ColTrack, it can be seen that the calculation overhead caused by the introduction of historical queries and several IRM modules is small. This is because these only slightly affect the calculation of the decoder part, and the calculations of CNN and encoders are not affected.

   \begin{table}
      \footnotesize
      \begin{center}
      \setlength{\tabcolsep}{1mm}{
      \begin{tabular}{lcccccccc}
      \hline
      \multirow{2}{*}{Method} & \multicolumn{2}{c}{MOT17} & \multicolumn{3}{c}{Dancetrack} & \multicolumn{3}{c}{BDD100K} \\
       \cmidrule(r){2-3}  \cmidrule(r){4-6}  \cmidrule(r){7-9} 
       & \(n\!\!=\!\!1\) & \(n\!\!=\!\!10\) & \(n\!\!=\!\!1\) & \(n\!\!=\!\!6\) & \(n\!\!=\!\!10\)  & \(n\!\!=\!\!1\) & \(n\!\!=\!\!6\) & \(n\!\!=\!\!10\) \\
      \hline
      & \multicolumn{8}{c}{IDF1}\\
      \hline
      \hline
      APPTracker               & 68.7 & 70.3         & - & - & -        & - & -  & -        \\
      BL+Bytetrack             & 77.5 & 64.4   & 47.1 & 31.6  & 27.3     & 41.3 & 22.6 & 20.3     \\
      BL+E2E                   & 77.6 & 74.8   & 50.6 & 55.5  & 49.9     & 51.0 & 50.7 & 47.4     \\
      \textbf{ColTrack}       & 78.1 & 78.6   & 54.6 & 61.6  & 51.3     & 54.0 & 52.7 & 51.4     \\
      \hline
      & \multicolumn{8}{c}{MOTA}\\
      \hline
      \hline
      APPTracker          & 68.7 & 65.5       & - & - & -            & - & - & -      \\
      BL+Bytetrack        & 75.3 & 61.0   & 89.4 & 72.2  & 62.3     & 29.4 & 14.6 & 13.2     \\
      BL+E2E              & 73.8 & 64.9   & 88.9 & 86.3  & 79.7     & 36.1 & 34.7 & 30.6     \\
      \textbf{ColTrack}  & 76.5 & 68.7   & 86.6 & 86.5  & 80.7     & 40.0 & 37.0 & 35.7     \\
      \hline
      & \multicolumn{8}{c}{HOTA} \\
      \hline
      \hline
      FairMOT                  & 56.9 & 51.1   & 37.6 & 26.3  & 23.4     & -    & -    & -        \\
      Bytetrack                & 67.5 & 57.6   & 46.1 & 32.3  & 29.4     & -    & -    & -        \\
      OC-SORT                  & 66.1 & 56.1   & 52.2 & 35.9  & 30.3     & -    & -    & -        \\
      MOTR                     & 62.1 & 60.0   & 51.7 & 52.2  & 47.3     & -    & -    & -        \\
      BL+Bytetrack             & 65.2 & 55.9   & 45.8 & 30.8  & 26.8     & 33.7 & 22.1 & 21.3     \\
      BL+E2E                   & 64.9 & 62.5   & 55.6 & 58.4  & 52.8     & 42.3 & 43.0 & 41.0     \\
      \textbf{ColTrack}       & 65.5 & 65.8   & 57.9 & 61.9  & 53.7     & 45.0 & 45.3 & 44.6     \\
      \hline
      \end{tabular}}
      \end{center}
      \caption{Performance comparison of different methods on videos at different frame rates on the validation set of three datasets. BL denotes Baseline. APPTracker~\cite{zhou2022apptracker} is an existing method devised for MOT under low frame rates.}
      \label{tab:abla:fr}
      \vspace{-0.02\linewidth}
      \end{table}

   \textbf{Verification under various scenarios}. 
   Further, in Table.~\ref{tab:abla:fr}, we compare the performances of different methods in videos at varied frame rates on three datasets. Compared with APPTracker\cite{zhou2022apptracker} devised for MOT in low-frame-rate videos, ColTrack achieves higher performance on both low-frame-rate videos as well as high-frame-rate videos. APPTracker pays more attention to the detection and association of emerging targets and does not fully solve the problems in the follow-up tracking of fast-moving targets. 
   
   Compared with Baseline+Bytetrack and Baseline+E2E, ColTrack has a significant performance improvement. The erratic movements and similar appearance of the dancers in Dancetrack make tracking difficult for the classical association method ByteTrack~\cite{zhang2022bytetrack} based on Kalman filtering. Meanwhile, the large driving dataset BDD100K with multiple categories contains more fast-moving objects. This makes object tracking very challenging in its low-frame-rate videos. Benefiting from the introduction of collaborative tracking queries, ColTrack still has satisfactory performance when \(n\) is varied from 1 to 10 on BDD100K.

   \subsection{Ablation Study}

   \begin{table}
      \footnotesize
      \begin{center}
         \setlength{\tabcolsep}{1mm}{
      \begin{tabular}{lcccccc}
      \hline
      \multirow{2}{*}{Method} & \multicolumn{3}{c}{MOT17} & \multicolumn{3}{c}{Dancetrack} \\
      \cmidrule(r){2-4}  \cmidrule(r){5-7}
      & HOTA & IDF1 & MOTA & HOTA & IDF1 & MOTA \\
      \hline

      BL+Bytetrack                  & 59.3 & 69.1 & 66.5 & 30.8 & 31.6 & 72.2  \\
      BE (BL+E2E)                   & 64.5 & 78.1 & 67.3 & 58.4 & 55.5 & 86.3  \\
      BE+CTQ                        & 65.1 & 78.5 & 70.4 & 59.3 & 57.1 & 86.1  \\
      BE+CTQ+IRM (rem)              & 66.1 & 79.2 & 71.7 & 60.0 & 57.3 & 86.0   \\
      BE+CTQ+IRM (add)              & 66.6 & 79.9 & 71.1 & 60.6 & 58.6 & 86.6   \\
      BE+CTQ+IRM                    & 66.7 & 80.0 & 72.2 & 61.0 & 58.9 & \textbf{87.6}   \\
      BE+CTQ+IRM+TOC                & \textbf{67.0} & \textbf{80.6} & \textbf{74.1} & \textbf{61.9} & \textbf{61.6} & 86.5   \\
      \hline
      \end{tabular}}
      \end{center}
      \caption{Ablation study of the components in ColTrack on the validation set of MOT17 and DanceTrack with downsampling interval \(n=6\). BL means Baseline. BE means BL+E2E.  CTQ denotes collaborative tracking queries. IRM is the information refinement module. IRM (add) means only the addition branch in IRM is kept. IRM (rem) means only the removal branch is kept. TOC denotes the tracking object consistency loss (TOCLoss). }
      \label{tab:abla:mot17}
      \vspace{-0.02\linewidth}
      \end{table}

   In Table.~\ref{tab:abla:mot17}, we analyze the impact of each module on the model performance under low frame rates (\(n=6\)). In BL+E2E+CTQ, although collaborative tracking queries are used, there is no interaction between historical queries of the same track. They can only assist the model training by interacting with queries of other tracks as negative samples through temporal blocking decoders. Therefore, the improvement BL+E2E+CTQ brings is very small. 
   
   When IRM is adopted, the performance of the model is significantly improved. This is because IRM enables collaborative tracking queries to interact with each other and refine themselves with temporal clues. We also analyze the impact of the information removal branch and the information addition branch in IRM on the model performance. Experimental results show that having both information removal and information addition capabilities helps the model refine features better.
   
   After adding TOCLoss, the performance of the model is further improved. TOCLoss forces collaborative tracking queries to refine themselves with IRM to better track corresponding targets. Then, they provide better temporal clues during iterative refinement in the following decoders. 

   In Fig.~\ref{fig:abla:method}(b) we analyze the effect of the max number of collaborative tracking queries \(N_{\mathrm{max}}\) on the tracking performance of ColTrack. The results indicate that ColTrack achieves best performances when \(N_{\mathrm{max}}\!=\!3\). A too large \(N_{\mathrm{max}}\) results in too many historical features being included, which introduces too much noise.
   
   \textbf{Analysis of the location and number of IRMs}.
   We also compare the performance of ColTrack with the different number of IRM modules in Fig.~\ref{fig:abla:method}(c). 6 decoders can add up to 6 IRM modules. We gradually remove the corresponding IRM in the shallow decoder layers. The experimental results indicate that removing the IRM before the first decoder layer and adding an IRM before each of the following decoder layers achieves the best performance. This is because queries have not interacted with the features of the current frame before passing through the first decoding layer, which makes tracking queries unable to adjust the fusion of temporal information according to the current frame. After the first decoder, more IRM modules allow queries to have more opportunities to communicate new observations with each other to gather more valuable information.

   \subsection{Comparison to the State-of-the-art Methods}

   \begin{table}
      \footnotesize
      \begin{center}
      \setlength{\tabcolsep}{1mm}{
      \begin{tabular}{lccccc}
      \hline
      Method & Source & HOTA & IDF1 & MOTA & IDs \\
      \hline
      TraDes~\cite{wu2021track}                           & CVPR'21 & 52.7 & 63.9 & 69.1 & 3555       \\
      FairMOT~\cite{zhang2021fairmot}                     & IJCV'21 & 59.3 & 72.3 & 73.7 & 3303        \\
      MTrack~\cite{yu2022towards}                         & CVPR'22 & -    & 73.5 & 72.1 & 2028        \\
      Unicorn~\cite{yan2022towards}                       & ECCV'22 & 61.7 & 75.5 & 77.2 & 5379         \\
      YOLOX+Bytetrack~\cite{zhang2022bytetrack}           & ECCV'22 & 63.1 & 77.3 & 80.3 & 2196         \\
      OC-SORT~\cite{cao2022observation}                   & CVPR'23 & \textbf{63.2} & 77.5 & 78.0 & 1950        \\
      P3AFormer (Swin)~\cite{zhao2022tracking}            & ECCV'22 & - & \textbf{78.1} & \textbf{81.2} & 1893    \\
      \hline
      \multicolumn{6}{c}{E2E methods}  \\
      \hline
      CenterTrack~\cite{zhou2020tracking}                  & ECCV'20 & 52.2 & 64.7 & 67.8 & 3039    \\
      Chained-tracker~\cite{peng2020chained}               & ECCV'20 & 49.0 & 57.4 & 66.6 & 5529    \\
      TransTrack~\cite{sun2020transtrack}                  & arXiv'20 & -    & 63.9 & 74.5 & 3663    \\
      TrackFormer~\cite{meinhardt2022trackformer}          & CVPR'22 & -    & 68.0 & 74.1 & 2829    \\
      MeMOT~\cite{cai2022memot}                            & CVPR'22 & 56.9 & 69.0 & 72.5 & 2724    \\
      MOTR~\cite{zeng2022motr}                             & ECCV'22 & 57.8 & 68.6 & 73.4 & 2439    \\
      \hline
      \textbf{ColTrack}                                    & - & 61.0 & 73.9 & 78.8 & \textbf{1881}    \\
      \hline
      \end{tabular}}
      \end{center}
      \caption{Performance comparison between ColTrack and the state-of-the-art methods under the ``private detector'' protocol on MOT17 test set.}
      \label{tab:test:mot}
      \end{table}

      \begin{table}
         \footnotesize
         \begin{center}
            \setlength{\tabcolsep}{1mm}{
         \begin{tabular}{lccccc}
         \hline
         Method & Source & HOTA & IDF1 & MOTA & AssA \\
         \hline
         FairMOT~\cite{zhang2021fairmot}              & IJCV'21 & 39.7 & 40.8 & 82.2  & 23.8    \\
         TransTrack~\cite{sun2020transtrack}          & arXiv'20 & 45.5 & 45.2 & 88.4  & 27.5    \\ 
         CenterTrack~\cite{zhou2020tracking}          & ECCV'20 & 41.8 & 35.7 & 86.8  & 22.6    \\    
         TraDes~\cite{wu2021track}                    & CVPR'21 & 43.3 & 41.2 & 86.2  & 25.4    \\
         QDTrack~\cite{pang2021quasi}                 & CVPR'21 & 54.2 & 50.4 & 87.7  & 36.8    \\
         YOLOX+Bytetrack~\cite{zhang2022bytetrack}    & ECCV'22 & 47.7 & 53.9 & 89.6  & 32.1    \\
         MOTR~\cite{zeng2022motr}                     & ECCV'22 & 54.2 & 51.5 & 79.7  & 40.2    \\
         OC-SORT~\cite{cao2022observation}            & CVPR'23 & 55.1 & 54.2 & 89.4  & 38.0    \\
         MOTRv2~\cite{zhang2022motrv2}                & CVPR'23 & 69.9 & 71.7 & 91.9  & 59.0    \\
         MOTRv2~\cite{zhang2022motrv2} (+val+ens)     & CVPR'23 & 73.4 & 76.0 & 92.1  & 64.4    \\
         \hline
         \textbf{ColTrack}                           & - & 72.6 & 74.0 & 92.1  & 62.3    \\
         \textbf{ColTrack} (+val)                    & - & \textbf{75.3} & \textbf{77.3} & \textbf{92.2}  & \textbf{66.9}    \\
         \hline
         \end{tabular}}
         \end{center}
         \caption{Evaluation results on the test set of Dancetrack. \emph{+val} means adding the validation set for training. \emph{+ens} denotes test ensemble.}
         \label{tab:test:dance}
         \vspace{-0.03\linewidth}
         \end{table}

         \begin{table}
            \footnotesize
            \begin{center}
               \setlength{\tabcolsep}{1mm}{
            \begin{tabular}{lccccc}
            \hline
            Method & Source & Split & mMOTA  & mIDF1 & IDs \\
            \hline
            Yu \etal ~\cite{yu2020BDD100K}                    & CVPR'20 & val  & 25.9 & 44.5   & 8315     \\
            DeepBlueAI ~\cite{deepblueai}                     & CVPRC'20 & val  & 26.9 & -      & 13366   \\
            QDTrack~\cite{pang2021quasi}                      & CVPR'21 & val  & 36.6 & 50.8    & 6262   \\
            MOTR~\cite{zeng2022motr}                          & ECCV'22 & val  & 32.0 & 43.5   & \textbf{3493}    \\
            Unicorn(ResNet)~\cite{yan2022towards}           & ECCV'22 & val  & 35.1 & -   & -    \\
            YOLOX+Bytetrack~\cite{zhang2022bytetrack}         & ECCV'22 & val  & 39.4 & 48.9  & 27902    \\
            \hline
            \textbf{ColTrack}                                & - & val  & \textbf{40.0} & \textbf{54.0}   & 3741   \\
            \hline
            Yu \etal ~\cite{yu2020BDD100K}                    & CVPR'20 & test  & 26.3 & 44.7   & 14674     \\
            DeepBlueAI ~\cite{deepblueai}                     & CVPRC'20 & test  & 31.6 & 38.7   & 25186   \\
            QDTrack~\cite{pang2021quasi}                      & CVPR'21 & test  & 35.5 & 52.3   & 10790    \\
            \hline
            \textbf{ColTrack}                                & - & test  & \textbf{40.4} & \textbf{56.0}  & \textbf{6249}    \\
            \hline
            \end{tabular}}
            \end{center}
            \caption{Comparison of the state-of-the-art methods on BDD100K.}
            \label{tab:test:bdd}
            \end{table}

      In Table.~\ref{tab:test:mot}, Table.~\ref{tab:test:dance} and Table.~\ref{tab:test:bdd}, we compare ColTrack with the state-of-the-art methods on three datasets. MOT17 is a small dataset containing only 7 training videos and 7 testing videos. Although such a small amount of data is difficult to train an end-to-end model to learn temporal relationship modeling, ColTrack still outperforms existing end-to-end methods and achieves comparable performance to existing tracking-by-detection methods. Dancetrack containing 100 videos is a large challenging dataset due to the irregular movements, similar clothing, and severe occlusions of the dancers. ColTrack outperforms all methods on Dancetrack. This is benefited by our introduction of collaborative tracking queries, which provide more abundant descriptions of targets and makes ColTrack less susceptible to similar appearances and occlusions. BDD100K containing 2000 driving videos is also a challenging dataset due to complex scenes and more fast-moving objects. ColTrack also achieves the best performance on BDD100K, especially for the IDF1 metric that focuses more on association performance. 

      These sufficient experimental results show that on more challenging datasets, our method not only achieves higher performance than existing methods under high frame rates but also better tracks objects in low-frame-rate videos. This fully proves that ColTrack is a frame-rate-insensitive model. It achieves faster processing speed while ensuring high performance at different frame rates.

\section{Conclusion}\label{sec:conclusion}

In this paper, we propose a collaborative tracking learning method (ColTrack) to address the challenges introduced by low frame rates in multi-object tracking (MOT). By introducing multiple historical queries to track the same target, rich temporal clues are used to obtain more comprehensive and accurate descriptions of the targets. We carefully devise the temporal blocking decoders and the information refinement module (IRM) such that the model allows collaborative tracking queries to better integrate the information while retaining the ability to inhibit duplicate predictions. Meanwhile, the proposed tracking object consistency loss (TOCLoss) forces each historical query to integrate valuable clues from other queries for the correct tracking. Thanks to the collaboration of these modules, ColTrack outperforms existing methods and achieves faster processing speeds on more challenging datasets Dancetrack and BDD100K at both high and low frame rates.

{\small
\bibliographystyle{ieee_fullname}
\bibliography{main}
}

\end{document}